%% file: main.tex
\documentclass[10pt,twocolumn,letterpaper]{article}

\usepackage{wacv}
\usepackage{times}
\usepackage{epsfig}
\usepackage{float}
\usepackage{graphicx}
\usepackage{amsmath}
\usepackage{amssymb}
\usepackage[accsupp]{axessibility}  
\usepackage{multirow}
\usepackage{adjustbox}
\usepackage{colortbl}
\usepackage{booktabs}
\usepackage[english]{babel}
\usepackage{lipsum}

\def\wacvPaperID{1224} 

\wacvfinalcopy 

\ifwacvfinal
\fi

\ifwacvfinal
\usepackage[breaklinks=true,bookmarks=false]{hyperref}
\else
\usepackage[pagebackref=true,breaklinks=true,colorlinks,bookmarks=false]{hyperref}
\fi

\begin{document}

\title{Challenges in Procedural Multimodal Machine Comprehension:\\A Novel Way To Benchmark}


\author{Pritish Sahu$^{1,2}$ \thanks{Work done while interning at SRI International.}   \thanks{These two authors contributed equally.} \quad
Karan Sikka$^{1}$\footnotemark[2]  \quad
Ajay Divakaran$^{1}$ \\
$^{1}$SRI International \\
$^{2}$Rutgers University \\
{\tt\small pritish.sahu@rutgers.edu, \{karan.sikka, ajay.divakaran\}@sri.com}
\\
}

\maketitle

\def\etal{et al\onedot}
\def\etc{etc\onedot}
\def\ie{i.e\onedot}
\def\eg{e.g\onedot}
\def\ck{Control-Knob{}}
\def\cks{Control-Knobs{}}
\def\vit{ViT}
\def\clip{CLIP}
\def\bidaf{BiDAF}

\def\m3c{$\text{M}^3\text{C}$}

\definecolor{redcol}{rgb}{1, 0, 0}
\definecolor{bluecol}{rgb}{0, 0, 1}
\newcommand{\red}[1]{\textcolor{redcol}{#1}} 
\newcommand{\blue}[1]{\textcolor{bluecol}{#1}} 
\renewcommand{\paragraph}[1]{\smallskip\noindent{\bf{#1}}}
\newcommand{\todo}[1]{\red{TODO: {#1}}}
\newcommand{\colons}[1]{``{#1}''}

\def\algorithmautorefname{Algorithm}
\def\figureautorefname{Figure}
\def\tableautorefname{Table}
\def\equationautorefname{Eq.}
\def\sectionautorefname{Section}

\ifwacvfinal
\thispagestyle{empty}
\fi

\ifx\compilemain\undefined

\input{abstract}
\input{intro2}
\input{related_work}

\input{approach}
\input{exp} 
\input{related_work} 
\input{conclusion}
\else


\fi

{\small
\bibliographystyle{ieee_fullname}
\bibliography{egbib}
}
\clearpage
\newpage
\appendix

\input{appendix}
\end{document}

%% file: abstract.tex
\begin{abstract}

We focus on Multimodal Machine Reading Comprehension
(M3C) where a model is expected to answer questions
based on given passage (or context), and the context and
the questions can be in different modalities. Previous works
such as RecipeQA have proposed datasets and cloze-style
tasks for evaluation. However, we identify three critical biases
stemming from the question-answer generation process
and memorization capabilities of large deep models. These
biases makes it easier for a model to overfit by relying on
spurious correlations or naive data patterns.
We propose a systematic framework to address
these biases through three Control-Knobs that enable us
to generate a test bed of datasets of progressive difficulty
levels. We believe that our benchmark (referred to as Meta-
RecipeQA) will provide, for the first time, a fine grained estimate of a model’s
generalization capabilities.
We also propose a general \m3c model that is used to realize
several prior SOTA models and  motivate a novel
hierarchical transformer based reasoning network (HTRN).
We perform a detailed evaluation of these models with different
language and visual features on our benchmark. We
observe a consistent improvement with HTRN over SOTA
($\sim 18\%$ in Visual Cloze task and $\sim 13\%$ in average over all
the tasks). We also observe a drop in performance across all
the models when testing on RecipeQA and proposed Meta--RecipeQA (e.g. $83.6\%$ versus $67.1\%$ for HTRN), which shows that the proposed dataset is relatively less biased. We conclude by highlighting the impact of the control knobs with some quantitative results.
\vspace{-0.8em}

\end{abstract}

%% file: intro2.tex

\section{Introduction}

\begin{figure}[htbp]
    \begin{center}
       \includegraphics[width=0.9\linewidth]{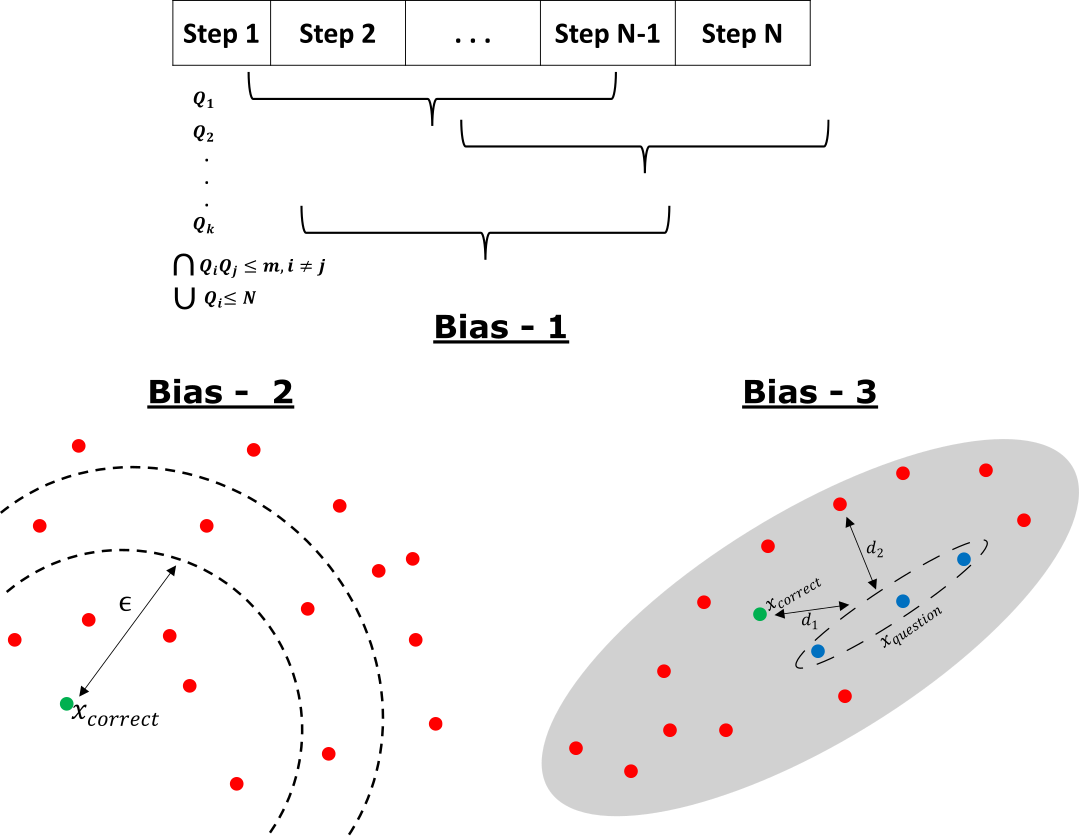}
    \end{center}
    \caption{An illustration of the three biases present in the dataset. \textbf{Bias-1} is caused when the overlap between the questions reveal the entire recipe process. The constraint is to upper bound the intersection of steps in multiple questions. \textbf{Bias-2} exists when the distance between the incorrect choices from the correct choice ($\epsilon$) is large in the latent space. The ``m" mentioned above is some small value. \textbf{Bias-3} occurs when the correct choice is closer to the question list ($|d_1-d_2|\leq\epsilon$) as compared to incorrect choices.}
    \label{fig:knobs}
    \vspace{-1.6em}
\end{figure}

Machine Reading Comprehension (MRC) has been used extensively to evaluate language understanding capabilities of Natural Language Processing (NLP) systems \cite{zeng2020survey, gardner2019making, richardson2013mctest, rajpurkar2016squad}. 
MRC is evaluated similar to how humans are evaluated for understanding a piece of text (referred to as \textit{context}) by asking them to answer questions about the text.  
Recently Multi-Modal Machine Comprehension (\m3c) has extended MRC by introducing multimodality in the context or the question or both \cite{sahu2021towards,kembhavi2016diagram,yagcioglu2018recipeqa,antol2015vqa} (\autoref{fig:model_diagram}). 
 A strong \m3c system is thus required to not only understand the (unimodal) context but also reason across different modalities.
Previous MRC studies have shown that it is often hard to to verify whether the model is actually understanding the context or naively using spurious correlations to answer questions \cite{gardner2019making, kaushik2018much, sahu2021towards}. 
We first identify three key biases that plague \m3c cloze-style benchmarks and then propose a novel procedure to create multiple datasets of different levels of difficulty from a single meta dataset.  
We then use these datasets to study the performance of different \m3c models and understand how the datasets affects performance. We also propose a novel hierarchical transformer based approach and show consistent improvements over prior methods.


\m3c can be evaluated in multiple ways \cite{zeng2020survey, liu2019neural, iyyer2017amazing, tapaswi2016movieqa, kembhavi2016diagram}. For example, in VQA the context is an image and the question and answer are in textual modality. 
\m3c datasets (\eg RecipeQA \cite{yagcioglu2018recipeqa}) can also include multiple modalities in the context or the question.
Multiple-choice cloze-style tasks are also used for evaluation, where the question is prepared as a sequence of steps with one of the steps replaced by a placeholder and the model is asked to find the correct answer from a set of choices.
Cloze-style evaluation is quite common in MRC since such questions can be generated without any human intervention. This makes it easier to train, test and deploy a model for a new domain with sparsely labeled data.
We focus on procedural \m3c, where the context is a list of steps, for preparing a recipe, that are described in multiple modalities.
We evaluate on the three visual cloze-style tasks defined in RecipeQA-- Visual Cloze, Visual Coherence, and Visual Ordering. 
Although several works have used RecipeQA for evaluation, we observe three critical biases introduced by 
how the cloze-style question-answers (QA) pairs were created for these tasks. These biases makes it easier for models to answer question by relying on spurious correlations and surface level patterns and thus casting doubts around prior evaluations.
The first bias is related to overlap between the questions, which are sampled randomly from four locations in the context in one of the modality.
This bias results from multiple questions being sampled from the same context and makes it easier for the model to answer questions by using other questions in the dataset.
The second bias results from the negative choices being far away from the correct choices \ie they are not hard negatives. Hard negatives refer to incorrect choices that are closer to the correct choice visually and might be difficult for a naive algorithm (\eg using background) to discriminate (\autoref{fig:oldvssnew}).  
This bias also causes the model to overfit as it can rely on simple features to get the correct answer.
The third bias is induced by the correct choice being significantly closer to the question (in feature space) as compared to the incorrect choices. This causes the model to answer questions by only matching the choices with the question. 
We propose a systematic way to tackle these biases by grounding them in three \textbf{\ck} that are then used to sample multiple datasets from a single meta dataset of recipes (\autoref{fig:knobs}). We refer to our benchmark as \textbf{Meta-RecipeQA}. 
We then use these knobs to generate datasets with lower bias and of progressively increasing difficulty level. For example, the accuracy of our model on the simplest and the hardest sets on Visual Cloze task varies from $56.3\%$ to $68.4\%$ on Meta-RecipeQA as compared to $70.5\%$ on RecipeQA.  
We also used additional pre-processing steps to improve the quality of the meta-dataset which is based on RecipeQA (\autoref{tab:meta_stats}).
We recommend that a model should be evaluated on all these datasets instead of a single sampled dataset to get a better estimate of its capabilities to answer questions. This type of evaluation is similar to cross-validation in machine learning which tends to provide improved estimates of performance.

In addition, we also study the effect of using better visual features for constructing these datasets and show that it has a large effect on performance. 
We also propose a general \m3c model (G\m3c) that is then used to realize several state-of-the-art (SOTA) models. 
G\m3c is composed of two primary components-- modality encoder and scoring function-- and allows us to systematically study the performance variations with different component choices.
We also got inspired by the general model to propose a \textbf{Hierarchical Transformer based Reasoning Network (HTRN)} that uses transformers for both the primary components. 
We show consistent improvement over SOTA methods with our approach ($+ 18\%$ in visual cloze task and $+ 13\%$ in average over all task). We also undertake an extensive ablation study to show the impact of visual features, textual features, and the \ck{} on performance. We see consistent drop in performance across all the models on Meta-RecipeQA as compared to RecipeQA showing that our approach is able to create harder datasets and addresses the underlying biases.
We finally provide qualitative analysis to show the effect of \ck{} on question-answer pairs. 
Our contributions are summarized as follows:
\begin{itemize}
    \item We identify and locate the origins of the three critical biases
that bedevil the RecipeQA dataset. We propose a systematic framework (referred to as Meta-RecipeQA) to address these biases through three \cks{}. The \cks{} makes it possible for us to generate a test bed of multiple datasets of progressive difficulty levels. \vspace{-0.5em}
    \item Propose a general \m3c (G\m3c) model that is used to implement several SOTA models and motivate a novel Hierarchical Transformer based Reasoning Network (HTRN). HTRN uses transformers for both modality encoder and the scoring function. \vspace{-0.5em}
    \item HTRN outperforms SOTA by $\sim +18\%$ in Visual Cloze task and $\sim +13\%$ (absolute) in average over all the tasks. \vspace{-0.5em}
    \item We observe a considerable drop in performance across all the models when testing on RecipeQA and the proposed Meta-RecipeQA (\eg $83.6\%$ versus $67.1\%$ for HTRN).
\end{itemize}

%% file: related_work.tex
\section{Related Works}

QA tasks have been a popular method for evaluating a model's reasoning skills in NLP. 
One of the earliest forms of question-answering task is Machine Reading Comprehension (MRC) \cite{hirschman1999deep} that involves a textual passage and QA pairs. The answer format can be in a cloze-form (fill in the blanks), or could include finding the answer inside the passage or generation  \cite{chen2016thorough, hill2016goldilocks, richardson2013mctest, rajpurkar2016squad}. 
A generalization of the MRC is a new task that employs multimodality in the context or QA and is referred to as MultiModal Machine Comprehension (\m3c). 
Several datasets have been proposed to evaluate \m3c, \eg COMICSQA~\cite{iyyer2017amazing}, TQA~ \cite{kembhavi2016diagram}, MoviesQA~\cite{tapaswi2016movieqa} and RecipeQA~\cite{yagcioglu2018recipeqa}.  
Although our work is closely related to \m3c, we tackle the task of procedural \m3c, \eg RecipeQA  where the context is procedural description of an event. RecipeQA~\cite{yagcioglu2018recipeqa} dataset provides ``How-To" steps to cook a recipe written by internet users. Solving procedural-\m3c requires understanding the entire temporal process along with tracking the state changes. Procedural \m3c is investigated in \cite{amac2019procedural} on RecipeQA by keeping track of state change of entities over the course. However, the method falls short in  aligning the different modalities.


Bias in a dataset could be referred to as a hidden artifact that allow a model to perform well on it without learning the intended reasoning skills. These artifacts in the form of spurious correlations or surface level patterns boost model performance to well beyond chance performance. Sometimes the cause of these biases are partial input data~\cite{gururangan2018annotation, poliaketal2018hypothesis} or high overlaps in the inputs~\cite{mccoyetal2019right, dasgupta2018evaluating}. These biases influence various other tasks as well such as argument reasoning~\cite{niven2019probing}, machine reading comprehension~\cite{kaushik2018much}, story cloze tests~\cite{schwartz2017effect, cai2017pay}. We have investigated three biases that plague the visual tasks in RecipeQA. One major bias occurs due to the high overlap of steps present in the question, this differs from \cite{mccoyetal2019right} as the later involves unimodal data and the overlap occurring in word embeddings. The next bias shown in \cite{schwartz2017effect} occurs due to  the differences in writing style in the text modality. RecipeQA also suffers from difference in style as a bias but in the visual domain. The bias due to difference in style in the visual domain is introduced due to the lack of necessary constraint while preparing the QA task.

%% file: approach.tex
\section{Approach}

We describe our approach by using visual cloze style tasks to demonstrate the efficacy of our benchmark.
We describe the three critical biases present in these tasks that prevents a comprehensive assessment of \m3c models. We simultaneously outline our proposed solution through \textbf{\cks} that are used to generate multiple datasets from a meta dataset. We contrast it with RecipeQA. Next, we describe of the general \m3c model which we use to realize many prior models and finally propose a novel method based on hierarchical transformers.

\subsection{Visual Cloze-Style Tasks}

\begin{figure*}[t]
    \begin{center}
        \includegraphics[width=0.9\linewidth]{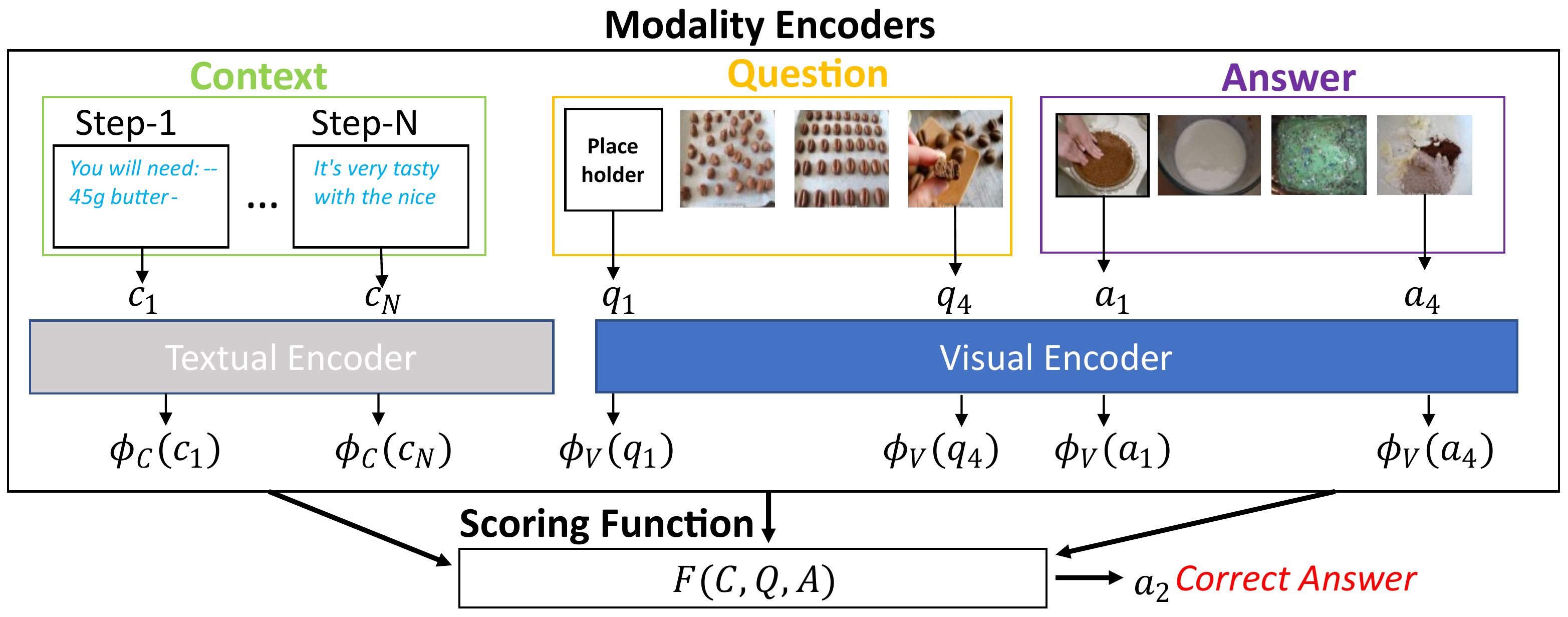}
    \end{center}
    \caption{An illustration of the General \m3c (G\m3c) model that consists of two primary components-- Modality Encoders and Scoring Function. We use this model to implement prior SOTA models but also propose our Hierarchical Transformer based Reasoning Network (HTRN) that uses transformers for both the components.}
    \label{fig:model_diagram}
\end{figure*}

We address the biases in the three visual cloze-style tasks from RecipeQA. In RecipeQA each instance (\autoref{fig:model_diagram}) is a sequence of steps about preparing a recipe, where each step is described by either images, text or both.
In the visual task, the context is in textual modality and the Question-Answer (QA) pairs\footnote{Here we refer to the sequence of steps with the placeholder as the question} are in visual modality.





%
\vspace{-0.5em}
\begin{enumerate}
    \item \textbf{Visual Cloze:} Determine the correct image that fits the placeholder in a question sequence of $N_Q$ images.
          The question is generated by selecting $N_Q$ images from a recipe and randomly replacing one of the images with a placeholder.
          \vspace{-0.5em}
    \item \textbf{Visual Coherence:} Determine the incoherent image from a list of $N_Q$ images. The coherent images are sampled in an ordered manner from one recipe.
          \vspace{-0.5em}
    \item \textbf{Visual Ordering:} Predict which sequence of images in the question is the correct sequence. The question is generated by sampling $N_Q$ images from $N_Q$ separate steps in a recipe and jumbling the order of the sequence for all except one.
\end{enumerate}

We set $N_Q$ to $4$ as done in RecipeQA.
In each of the tasks the model is expected to establish cross-modal correspondences across textual steps and visual QA pairs and then reason to find the correct answer. We selected these tasks as they cover a broad range of reasoning capabilities and also allows us to verify our approach for removing biases on multiple tasks.
The above mentioned skills are required to solve cloze style, coherence and ordering. \m3c cloze style, coherence and ordering skills assess the knowledge and understanding obtained from the context and question \cite{sahu2021comprehension}.

\begin{figure}[t]
    \begin{center}
        \includegraphics[width=0.9\linewidth]{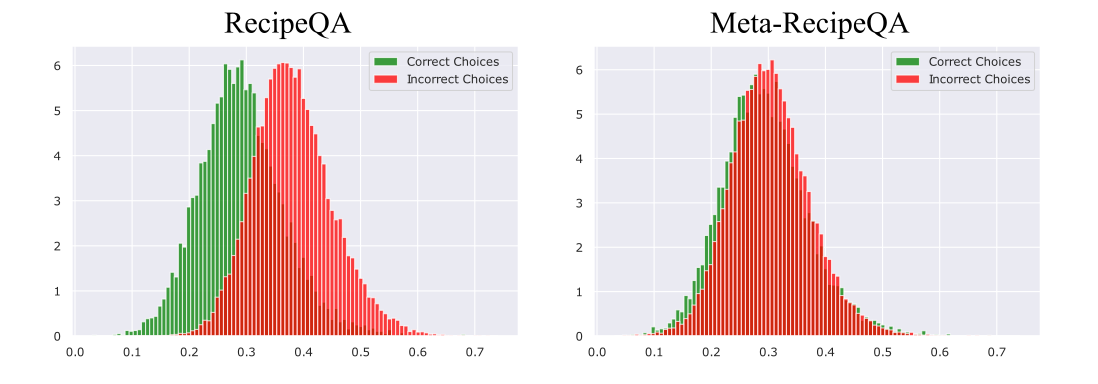}
    \end{center}
    \caption{Distribution of distances of the correct and incorrect choices from the question in the feature space (\vit{}) for RecipeQA (left) and one of our datasets generated with \ck{} set to (0,1,1). Training a Support Vector Machine (SVM) with the distance of each choice from the question as input feature results in an accuracy of $71.9 \%$ on RecipeQA and $31.7 \%$ on our dataset. This highlights the inherent bias in RecipeQA which can be exploited by a naive model to give high performance.}

    \label{fig:choice_dist}
    \vspace{-1.5em}
\end{figure}

\subsection{Biases and Proposed \cks}
\label{sec:knobs}

\m3c models are evaluated to measure whether they are able to understand the context and then answer questions.
However, it has been shown before that it is common for MRC models to answer questions by using biases or surface level patterns in the data \cite{sahu2021towards, gardner2019making,chen2016thorough}.
Such behavior stems both from the data creation process and large capacity of recent state-of-the-art (SOTA) models-- which makes it easier for them to overfit \cite{brown2020language}.

\autoref{fig:choice_dist} depicts an example of data bias in RecipeQA by showing the distribution of distances between the question (averaged) and the correct and incorrect choices in the feature space. We see that
the distributions are well separated in the original RecipeQA as compared to one of our datasets. This allows a model to answer questions without reading the context. With these inherent biases can prior evaluation results be trusted and how can we do better? We formalize three key biases present in previous evaluation setup and also propose a solution to counter them with \cks. We use the visual cloze task as an example to describe the \cks. We provide descriptions for the other two tasks (visual coherence and visual ordering) in Appendix B. We briefly describe the construction of QA pairs in visual cloze (from RecipeQA) to better understand our \cks.
Questions are prepared by repeating the process of first sampling four random locations (in increasing order) in the recipe and then replacing one of the images randomly with the placeholder.
The negative choices are randomly sampled, beyond a certain distance from the positive choice, in the feature space. We have illustrated these biases visually in \autoref{fig:knobs}.
\vspace{-0.5em}
\begin{enumerate}

    \item \paragraph{Bias-1-High overlap between question sequences}: This bias occurs since multiple question sequences are sampled from the same recipe. Here the model can learn the correct answer by relying on other oversampled questions and fail to actually understand the context.

          \paragraph{\ck-1:} This knob controls the overlap between the questions as well as the maximum number of questions that can be generated from a recipe. It first
          imposes a constraint on the maximum number of questions that can be sampled from a recipe. We also sample questions from recipes with $\#Steps\geq 5$. Although we iteratively sample a question from a recipe as done in RecipeQA, we minimize the overlap between questions by removing the step corresponding to the correct choice before sampling the next question. This makes sure that the model cannot exploit commonalities between questions to know the correct answer. We use two settings for this knob where the first setting fixes the maximum number of questions to $\#steps/2$. The second setting makes the dataset harder by fixing the maximum number of questions to $\#steps/3$ and also removes a random choice along with the correct choice before sampling the next question.





    \item \paragraph{Bias-2-Incorrect choices are not hard negatives}: This bias occurs when the correct choice is closer (in feature space) to the question features as compared to the incorrect choices. A model can thus exploit this artifact to answer question.


          \paragraph{\ck-2:} We first compute K nearest-neighbors (KNNs) of the correct choice and select $K_C$ points in the feature space. To vary the difficulty level of the negative choices, we discretize the space of KNNs by computing mean ($m_d$) and standard deviation ($\sigma_d$) of the distances of the $K_C$ points from the correct choice.
          We use two settings of this knob by either sampling the negative choice from the euclidean ball $(0, m_d - s_d)$ or $(m_d - s_d, m_d + s_d)$. The first setting with generate harder negatives since they will be closer to the correct choice.
          We use image features from pre-trained models for computing these distance and observe that such features have a huge impact on the semantic similarity of the incorrect choices to the correct choice (ViT versus ResNet-50).



    \item \paragraph{Bias-3-Incorrect choices being far away from the question:} This bias occurs since the correct choice is closer (in feature space) to the question features as compared to the incorrect choices. In such cases the model can simply answer question by using the relative distances between correct and incorrect choice to the question and can bypass the context. This is similar to using odd-one-out in standard comprehension.

          \paragraph{\ck-3:} Generally all the images from one recipe exhibit underlying semantic similarity such as the background. The incorrect choice should share some semantic similarities with question images, similar to the correct choice, to make it harder for the model to discriminate based on such naive cues. The \ck{} is designed to consider the distance between the the question and the correct choice when sampling the incorrect choices. During the process of selecting negative choices in \ck-2, we select one negative choice which is close to the question as compared to the correct choice. We use two values for this knob-- when this knob is off, we do not enforce the constraint described above; when the knob is on, we randomly select one negative choice to satisfy the distance constraint.


\end{enumerate}

\subsection{Meta Dataset for Meta-RecipeQA}

We refer to our proposed benchmark that consists of multiple datasets generated by varying the \cks{} as \textbf{Meta-RecipeQA}. We create these datasets using a meta dataset which contains
all the recipe from RecipeQA without any of the tasks. We found that several recipes in RecipeQA had missing/partial content and were noisy (\eg multiple words were joined together). We used its source (\url{instructables.com}) to complete and clean the existing content.
With our proposed Control-Knobs we are able to regulate the amount of question-answer generated from 8K to 22K. We also notice that RecipeQA dataset is plagued with out-of-vocabulary tokens ($19\%$ over all tokens). One main reason for out-of-vocabulary tokens is the fusion of multiple
in-vocabulary tokens. In the Meta Dataset, our cleaning process results in $90\%$ in-vocabulary tokens over all tokens. Please refer to the supplement for a detailed description involved in the process of creating the Meta Dataset.

\begin{table}[htbp]
    \centering
    \caption{\hspace*{0mm} Dataset statistics for RecipeQA and Meta-RecipeQA.  \label{tab:meta_stats}}
    \begin{adjustbox}{width=0.49\textwidth}
        \small
        \begin{tabular}{lcc}
            \toprule
                                               & \textbf{RecipeQA} & \textbf{Meta-RecipeQA} \\ \hline
            \#Recipes(train, valid )           & 9101              & 8639                   \\
            \#VisualCloze QA                   & 7986              & (8K-22K)               \\
            \#Recipes used in VisualCloze QA   & 5684              & (6K-9K)                \\
            \#in-vocab tokens / \#vocab tokens & $19.9\%$          & $90.2\%$               \\
            \bottomrule
        \end{tabular}
    \end{adjustbox}
    \label{table:meta_stats}
    \vspace{-1.3em}
\end{table}

\subsection{Hierarchical Transformer based Reasoning Network (HTRN)}

We now describe the general \m3c model (G\m3c), which is used to realize prior SOTA methods along with our proposed method. For description of this model we limit ourselves to the visual cloze task and provide additional details in Appendix C.
The context $C=\{c_k\}_{k=1}^{N_C}$ consists of $N_C$ steps in textual modality, where each step $c_k=\{w_s^k\}_{s=1}^{K}$ contains $K$ tokens.
For the visual cloze task, the question $Q=\{q_i\}_{i=1}^{N_Q}$ consists of $N_Q$ images with one image being replaced by a placeholder.
The answer $A=\{a_j\}_{j=1}^{N_A}$ is composed of one correct and $N_A - 1$ incorrect choices.

\paragraph{G\m3c:} is shown in \autoref{fig:model_diagram} and consists of two primary modules-- \textbf{modality encoder} and \textbf{scoring function}. The modality encoder featurizes the textual context using a textual encoder, denoted as $\phi_T(c_k)$ as well as the questions and the answers using a visual encoder, denoted as $\phi_V(q_i)$. The output from both these modules is fed into the scoring function to compute a compatibility scores for the answer.
We use this model to implement prior SOTA models.
For example, a popular method ``Impatient Reader" uses Doc2Vec with an LSTM as the textual encoder, ResNet50 with an LSTM as the visual encoder, and then uses attention layers for the scoring function.

\paragraph{Hierarchical Transformer based Reasoning Network (HTRN):} is build upon the G\m3c model.
HTRN encodes each step in the context using a pre-trained transformer model to obtain embeddings for each token. We also use a bi-directional LSTM to encode the contextual features for each step.
We obtain the feature vector for each step by averaging feature of all the tokens for that step.
To model temporal dependencies across the steps, HTRN uses another bi-directional LSTM before feeding the inputs to the scoring function.
For the visual encoder, we use the pre-trained transformer based visual encoder.
We now have the encoding for each step, question and answer as $\phi_C(c_k)$, $\phi_V(q_i)$, and $\phi_V(a_j)$ respectively. We also use a bi-directional LSTM to encode the temporal relationships between the images. These inputs are now passed to the scoring function.

The aim of the scoring function is to provide a score for each of the candidate answer $a_j$. Since we need to score $N_A$ answers, we create $N_A$ query vectors (denoted as $u$), where each query vector is prepared by replacing the placeholder with the candidate choice at location $j$ in the answer.
HTRN uses a second shallow transformer (trained from scratch) for the scoring function.
We use ideas from preparing BERT inputs for question-answering \cite{devlinetal2019bert} by creating a representation for a context-query pair as $R(C, a_j) = [\text{CLS}, \phi_C(c_1), \dots, \phi_C(c_N), \text{SEP},\phi_V(q_i), u$], where CLS and SEP are special token as used in the NLP models and $[]$ denotes concatenation.
We pass this input through the transformer and use the contextual representation of the CLS token as the final representation of $j^{th}$ query vector.
We finally use an FC layer to obtain scores for all the query vectors.
Our motivation for using the transformer as a scoring function is that its underlying self-attention mechanism enables us to model the complex relationships between the context-context, context-QA, and QA-QA pairs. Such relationships are often modeled by multiple components in prior models and may not suffice for the application at hand. Moreover, transformers bring additional advantage in terms of multi-head attention and skip-connection leading to improved learning \cite{devlinetal2019bert,radford2021learning}.

%% file: exp.tex
\section{Experiments}

\begin{figure*}[htbp]
    \begin{center}
        \includegraphics[width=0.99\linewidth]{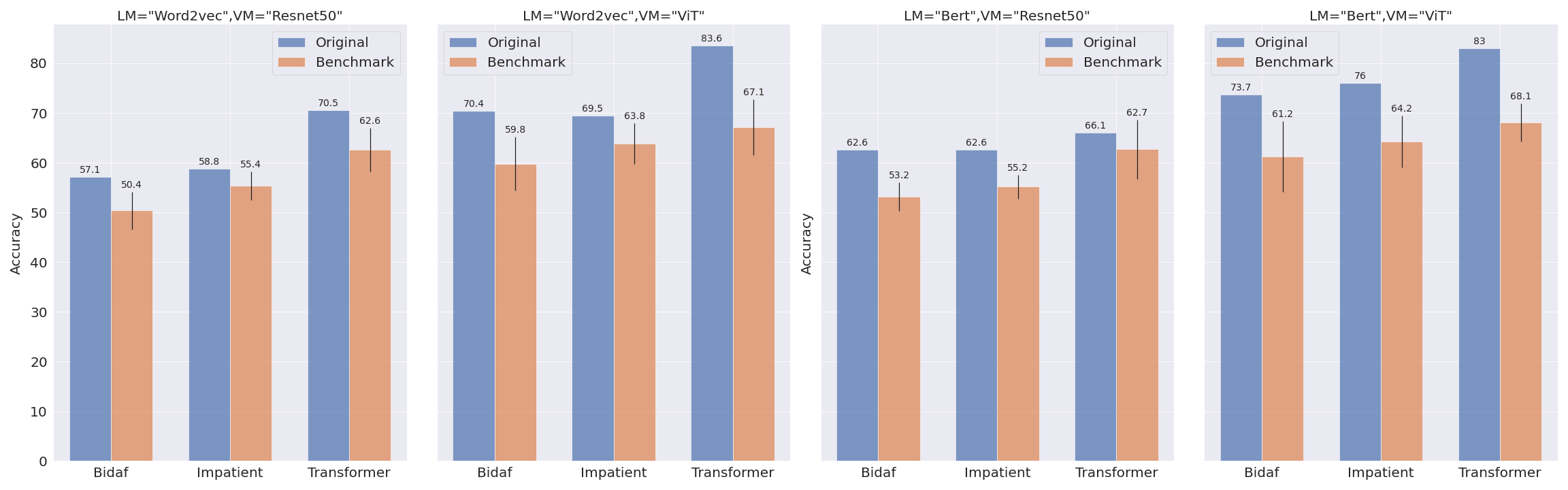}
    \end{center}
    \caption{Evaluation on visual cloze tasks on both RecipeQA (blue bar) and the proposed Meta-RecipeQA (yellow bar). For Meta-RecipeQA we compute the mean and the standard deviation of performance across all the generated datasets. For each figure, we mention the LM and VM used for training the model}
    \label{fig:oldvssnew}
\end{figure*}

In this section we empirically study the (1) impact of the \cks{} on the performance of different models, and (2) performance improvement from our proposed HTRN model.
We begin by stating the dataset creation process using the \cks{} as well the metrics used for evaluation, details of prior methods and the implementation details of our methods.
Next, we report the quantitative results where we first compare different models on RecipeQA with our proposed Meta-RecipeQA benchmark. Next, we study the impact of the \cks{} in more details.
We then compare the proposed HTRN models with SOTA methods.
We finally provide quantitative results to highlight the effect of the \cks{} on some generated question-answer pairs.


\subsection{Dataset and Metrics}

We use the 3 \cks{} to create multiple datasets for evaluation. To keep the number of experiments under control we use two discrete settings for each of these \ck{} (see in \autoref{sec:knobs}).
In the remainder of the text we shall refer to the dataset setting with the \cks{} set to value $i, j, k$ as a tuple $(i, j, k)$.
Along with the \cks{}, we use two choices of language models (LM) (Word2Vec, BERT), two choices of visual models (VM) (ResNet-50, ViT) and three different scoring function. For the visual cloze task, we trained a total of $108$ models. Out of $108$, $96$ models are trained on Meta-RecipeQA that constituted combinations of $3$ \cks{}, $2$ LM, $2$ VM and $3$ scoring functions.


We use classification accuracy as the metric that measures the percentage of questions that the model is able to answer correctly.

\vspace{-0.3em}
\subsection{Prior Methods and Implementation Details}
\vspace{-0.3em}
\paragraph{Prior Methods}: For comparison with SOTA we compare our model with Hasty Reader \cite{yagcioglu2018recipeqa}, ``Impatient Reader"\cite{hermann2015teaching}, PRN\cite{amac2019procedural}, MLMM-Trans \cite{liu2020multi} on RecipeQA. We obtain their results from MLMM-Trans \cite{liu2020multi}.
For the experiments involving \cks{} on proposed benchmark, we adopted two popular MRC models as the scoring function in addition to transformers in HTRN -- \bidaf \cite{seo2016bidirectional}, ``Impatient Reader" \cite{hermann2015teaching}.

\subsection{Comparison on RecipeQA and the Proposed Meta-RecipeQA}

In \autoref{fig:oldvssnew}, we show results of different algorithms with different visual and textual features on RecipeQA and Meta-RecipeQA. We report mean performance across our proposed datasets that were generated by sweeping through eight combination of the control knobs.

We first observe a consistent drop in performance, for all algorithms, between the previous and the proposed splits. For example, with LM as Word2Vec and VM as ViT, the performance of HTRN on old and proposed splits is $83.6\%$ and $67.1\%$ respectively. We also observe a similar drop with prior methods \eg for \bidaf{} the performance on RecipeQA and Meta-RecipeQA is $70.4\%$ and $59.8\%$ respectively. We believe this drop occurs since the \cks{} are able to remove some of the biases present in RecipeQA by creating a benchmark which makes it for the models to overfit. We also observe that the visual features have a large impact on performance \eg HTRN gives $67.1\%$ and $62.1\%$ with \vit{} and Resnet50 respectively. We also believe that it is easier for improved image features to overfit for the visual cloze task since they can easily compare two images using surface level patterns such as background. Also, the variance in performance across different splits highlights that our \ck{} provide flexibility in creating datasets with progressive difficulty levels (in terms of skills required for solving these tasks).




\begin{table}[htbp]
    \centering
    \caption{\hspace*{0mm}Comparison of \textbf{HTRN} with SOTA on the three visual tasks of RecipeQA dataset.\label{tab:HTRNresults}}
    \begin{adjustbox}{width=0.5\textwidth}
        \label{}
        \small

        \begin{tabular}{lllll}
            \toprule
            Model                                       & \textbf{Cloze} & \textbf{Coherence} & \textbf{Ordering} & \textbf{Average} \\ \hline
            Human* \cite{yagcioglu2018recipeqa}         & 77.6           & 81.6               & 64.0              & 74.4             \\ \hline
            Hasty Student \cite{yagcioglu2018recipeqa}  & 27.3           & 65.8               & 40.9              & 44.7             \\
            Impatient Reader \cite{hermann2015teaching} & 27.3           & 28.1               & 26.7              & 27.4             \\
            PRN \cite{amac2019procedural}               & 56.3           & 53.6               & 62.8              & 57.6             \\
            MLMM-Trans \cite{liu2020multi}              & 65.6           & 67.3               & 63.8              & 65.6             \\
            \midrule
            (Word2Vec, Resnet--50)                      &                &                    &                   &                  \\
            \midrule
            HTRN-Bidaf*                                 & 57.1           & 58.2               & 65.5              & 60.3             \\
            HTRN-Impatient*                             & 58.8           & 57.9               & 64.2              & 60.3             \\
            HTRN-Transformer*                           & 70.5           & 67.7               & 65.1              & 67.8             \\
            \midrule
            (BERT, ViT)                                 &                &                    &                   &                  \\
            \midrule
            HTRN-Bidaf                                  & 73.7           & 77.0               & 70.7              & 73.8             \\
            HTRN-Impatient                              & 76.0           & 74.1               & 70.7              & 73.6             \\
            HTRN-Transformer                            & \textbf{83.6}  & \textbf{80.1}      & \textbf{70.3}     & \textbf{78.0}    \\ \hline
        \end{tabular}
    \end{adjustbox}
    \label{table:sota}
    \vspace{-1.3em}
\end{table}

\subsection{Impact of Control Knobs on Performance}

\begin{figure*}[htbp]
    \begin{center}
        \includegraphics[width=0.99\linewidth]{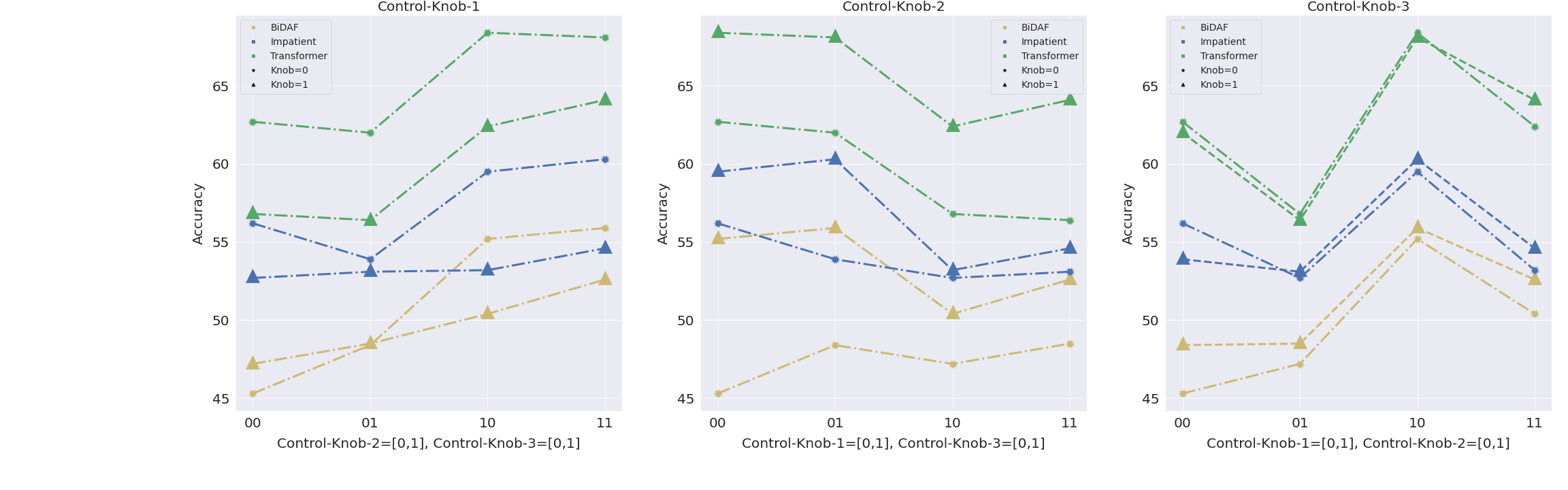}
    \end{center}
    \caption{Studying the impact of \cks{} by comparing the performance of different scoring functions (adapted baseline and transformer) for each knob setting. LM is set to Word2Vec and VM is set to ResNet--50. Starting from left we plot performance of \ck{-2} and \ck{-3} for all combination of control setting by fixing \ck{-1}. We do the same for \ck{-2} and \ck{-3} in the center and right figure respectively.}
    \label{fig:impact_knobs}
    \vspace{-1.3em}
\end{figure*}

In \autoref{fig:impact_knobs} we show the impact of different \ck{} on performance.
We measure performances for three models with LM and VM as Word2Vec and Resnet50 respectively on datasets generated by sweeping across the two discrete values for each of the \ck.
All the models achieved their best performance when \ck-1 (overlap bias) was set to $0$ i.e. high overlap and \ck-2 (distance of incorrect choice from correct choice) was set to $1$ i.e. images sampled in the euclidean ball between [$\mu_d-\sigma_d$, $\mu_d+\sigma_d$].
However, if we choose complement of these values we obtain lower performance across all the models. This is the case since reducing the overlap between questions and reducing the distance between correct and incorrect choices makes the QA harder. This highlights the ability of our \ck{} to create datasets with progressive difficulty levels.
For \ck-3, we see a small influence on model's performance. We believe this results from our implementation of \ck-3, where we randomly flip a coin on each sample and apply the constraint to only one incorrect choice. This \ck{} can be more impactful by applying it over multiple choices. The experimental details of HTRN is presented in Appendix A.

\subsection{Comparison with SOTA}

\autoref{table:sota} shows the comparison of our model HTRN with SOTA models on the three tasks from RecipeQA. We establish superiority by comparing our results against previous RecipeQA benchmark. We start by demonstrating the human performance on RecipeQA\cite{yagcioglu2018recipeqa} followed by results from prior works. Our model outperforms MLMM-Trans (multi level transformer based model) on all tasks. For example, the performance of HTRN versus MLMM on the ordering task is $65.1$ and $63.8$ respectively. We also believe that our algorithm is the first to outperform human's performance on the ordering task. We compare with the prior SOTA models using LM: Word2Vec and VM: ResNet-50 (under same input conditions), represented in the table \ref{table:sota} as HTRN-Transformer$^*$, where we get a $5\%$ and $2\%$ improvement over MLMM-Trans in visual cloze and ordering task respectively. With the last three rows, report the performance of our proposed model HTRN (LM: BERT,VM :ViT), where gain over MLMM-Trans by a margin of $12.4\%$ in average ($18\%$ in visual cloze, $13\%$ in coherence and $16.5\%$ in ordering). We also compared our transformer scoring function against Bidaf and Impatient Reader as scoring function. We observe improvements of $7.5\%$ (absolute) highlighting the performance advantage of the transformer based scoring function which is able to better model the complex relationships between multimodal context and QA pairs.


\subsection{Qualitative Analysis}
\begin{figure}[htbp]
    \centering
    \includegraphics[width=0.99\linewidth]{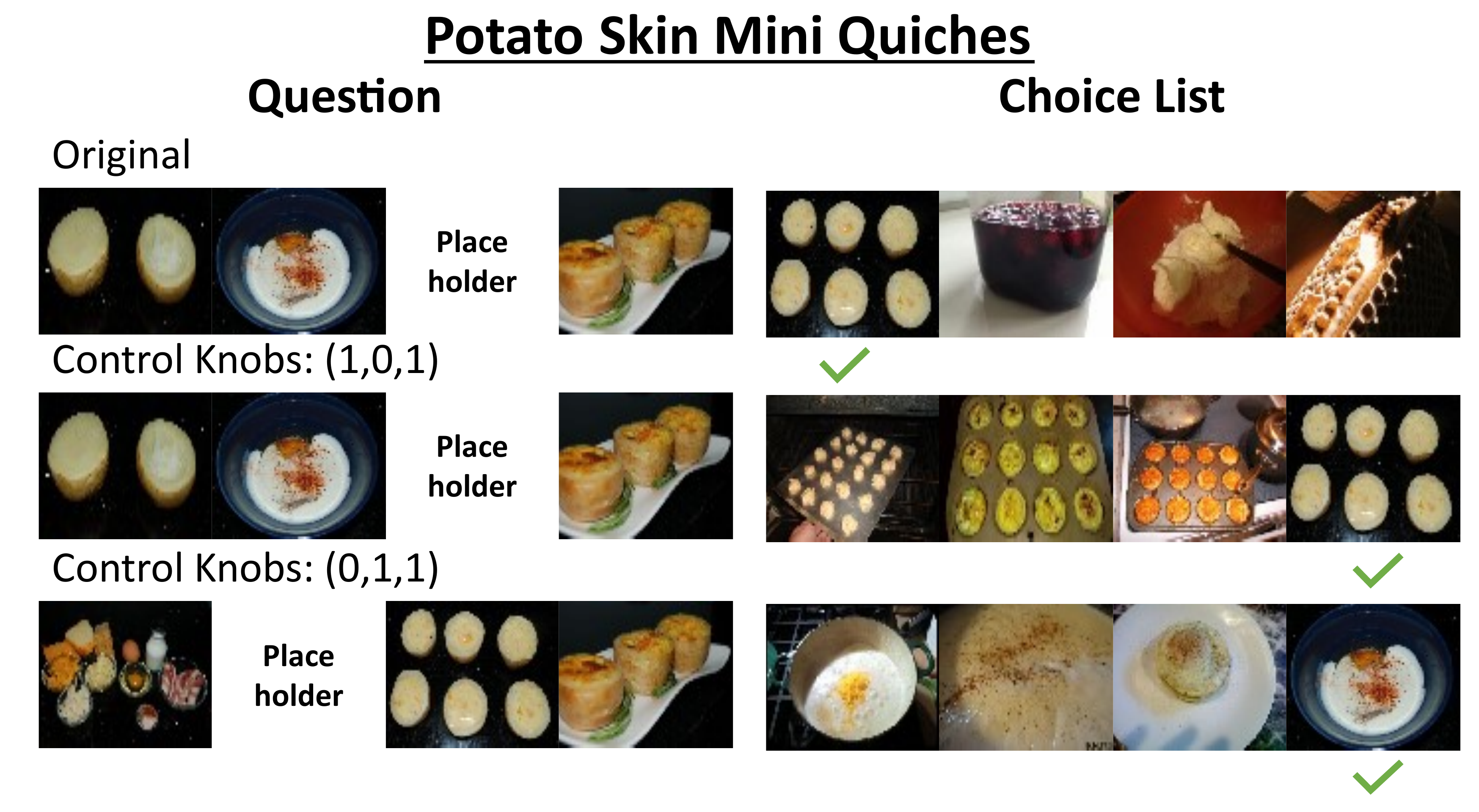}
    \caption{Question-Answer (QA) pairs generated using our \cks{} for recipe ``Potato Skin Mini Quiches". First row shows QA from RecipeQA. Row 2 uses \ck{} setting (1,0,1), best performance dataset on HTRN. Row 3 uses \ck{} setting (0,1,1) (from the dataset with the lowest performance). Distinguishing the correct images is easiest in the first QA pair compared to the second and third rows where two of the images share a similar background.}
    \label{fig:potato}
\end{figure}

\vspace{-0.5em}
In \autoref{fig:potato}, we show three questions from the same recipe with the choice list. The QA pair in the first row is chosen from the original RecipeQA dataset. The QA pairs in second and third rows are selected with Control Knobs set to (1, 0, 1) and (0, 1, 1)  to provide a medium and a hard QA pair. We can see that in the case of the first QA pair it will be easier for a model to use the background to find the correct answer.
However, this is slightly harder in the second row where two of the images share a similar background. We believe it would be hardest for a model to find the correct answer by using surface level patterns in the third QA pair, where the negative choices are semantically closer (in foreground) to the correct answer. We believe that such hard negatives will generally make it harder for a model to utilize the biases and thus help in generalization.

%% file: conclusion.tex
\section{Conclusion}

In this paper, we identified three key weaknesses
in the \m3c based RecipeQA dataset stemming from distance between question-choices, as well as overlap between questions. We propose a novel Meta-RecipeQA framework guided by three \cks{} to reduce
the bias in the question-answering tasks. Using the defined \cks{}, we propose multiple datasets of progressive difficulty levels. We also propose a general
\m3c (G\m3c) framework for realizing SOTA models. We use the
general \m3c to implement SOTA models such as BiDAF and
Impatient Reader. It also motivates us to propose a novel Hierarchical
Transformer based Reasoning Network (HTRN) that
uses transformers for both the modality encoders and the
scoring function. We significantly outperform prior SOTA methods
on RecipeQA. Similarly we see a drop in performance
over Meta-RecipeQA as compared to RecipeQA that
suggests that we have successfully reduced the bias. We
also gain deeper insights into each \cks{} through quantitative analysis.
We hope that our framework will provide a rich evaluation of multimodal comprehension systems by testing models on datasets generated by sweeping through the three proposed \cks.
Such evaluation will go beyond overall accuracy to a fine-grained understanding of robustness to dataset bias.

\textbf{Acknowledgement: } We would like to thank Michael Cogswell for the valuable suggestions. 

%% file: appendix.tex
\section{Implementation Details for HTRN}
In our experiments, we experiment using Word2Vec \cite{mikolov2013efficient} (trained on the step description of the recipes) and BERT model-- ``bert-basenli-mean-tokens" trained using SentenceTransformers \cite{reimers-2019-sentence-bert}. For visual encoder, we use the outputs of the final activation layer for ResNet50 trained on ImageNet dataset and \vit{} model \cite{radford2021learning} pretained with \clip{} \cite{radford2021learning}.
In our implementation, we used two single layer Bi-Directional LSTMs (size of the hidden layer is $256$) to model the temporal information within a recipe steps and across the steps.
The transformer based scoring function in HTRN was trained from scratch. This transformer uses 4 hidden layers with a size of $512$ and $8$ attention heads.
We used CrossEntropy loss to train our model. We use the Adam \cite{kingma2014adam} optimizer with the fixed  learning rate of $5\times10^{-4}$ including an early stopping criteria with a patience set to $20$ for all of our experiments. We performed our experiments on a single Nvidia GTX 1080Ti and that takes about $8-10$ hours to train for a single experiment. We did not perform any hyperparameter tuning and used the same hyperparameters for all the models trained.

\section{\cks{} Design Process}
As described in Section 2.2, question preparation begins by sampling four random locations (increasing order) in the recipe. We give details of the \cks{} that are unique to each of the remaining visual tasks i.e. visual coherence and ordering. \ck{-1} is applied to all the visual tasks.

\subsection{Visual Coherence}
\paragraph{\ck{-2}: }
In contrast to visual cloze where three negative choices are also provided along with the correct answer in the choice list, visual coherence on the other hand has only one incoherent image (correct answer) among the four images. For visual cloze, the selection of a negative choice was determined by it's metric distance from the correct choice whereas in coherence there are not multiple negative choices \ie the odd the image in the set is the correct answer. Hence, we alter \ck{-2} process to select the incoherent image from the set of union/intersection of K nearest neighbor (KNNs) of these three coherent images. We use two \ck{} setting and use the union set of the KNNs where the sampling space of negative choice is the Euclidean ball (0, $m_d - s_d$) or ($m_d - s_d$, $m_d + s_d$). However, the $m_d$ and $s_d$ are computed on the samples from the union set.

\paragraph{\ck{-3}}
As described in the main paper, the aim of this \ck{} is to control the distance of negative choice from the question. In case of visual coherence, we use the three coherent images as the question and the incoherent image as the negative choice. We take the min  of all pairwise distance for the coherent images. To make the negative choice closer, the distance of a randomly picked sample from the mean of question should be smaller than the min distance of all pairwise computed earlier. This forces
sample of a negative choice with similar features as the question.

\subsection{Visual Ordering}
\paragraph{\ck{-2}}
For a sequence of four images randomly selected (in increasing order), excluding the correct order, there is 23 other ordering present. The altering process for \ck{-2} involves devising a metric to control the sampling of three negative choices out of 23 possible choices. The metric devised is as follows, for each wrong sequence, we get a score by computing the pairwise distance of the consecutive images in the sequence and adding them \ie $\sum_{i=1}^{i=3}dist(\phi_V(x_i), \phi_V(x_{i+1}))$, where $x_i$ represents the image at $i^{th}$ index in the wrong sequence. We obtain 23 such scores which we refer to as weights to compute weighted probability distribution. We follow by associating each wrong sequence with their corresponding probability. We finally sample three negative choice out of 23 options based on the probability associated with them. We have two setting for this \ck{}, case-1 when the probability distribution is uniform, case-2 as discussed above. As the
visual ordering task evaluates the memorization and recall
ability of the model , we do not have \ck{-3}.

\section{Hierarchical Transformer based Reasoning Network (HTRN)}
In this section, we briefly describe the process of preparing query vectors for visual coherence and ordering tasks.

\subsection{Visual Coherence}
For Visual Coherence, the question $Q = \{q_i\}_{i=1}^N$ consists of $N_q$ images, including an incoherent image in between. The answer A is a scalar pointing to the location of the incoherent image. As we need $N_A$ scores for each candidate answer $a_j$, we create $N_A$ query vectors, where each query vector is prepared by removing one element from Q. Finally we obtain $N_A$ query vectors each of size $N_A - 1$ and only one vector contains all the coherent images.

\subsection{Visual Ordering}
As the structure of ordering task provides $N_A$ choice vectors each of size $N_A$. We do not make any changes here.


\section{Meta Dataset of Meta-RecipeQA}
The preparation of metadata is broken into two folds: 1) description of each recipe which is compiled from \cite{yagcioglu2018recipeqa} and \url{instructables.com}, 2) control panels that moderates the scale of the question and answer during the dataset generation process. Below, we describe the process involved in the creation of Meta Dataset.

Each recipe in the metadata stage contains the following information a) name of the recipe, b) all steps required for completing the recipe, where each step includes multi-modal (text, image) information describing that step. In order to prepare the meta recipe content, we begin by further cleaning the RecipeQA \cite{yagcioglu2018recipeqa} as we observe noise in each modality i.e. text and image. The missing texts and images from recipe steps were scrapped again from \url{instructables.com}. Next, we remove the noise in the textual side by processing the data again. Few of the persisting noise removed by our algorithm are "HTML/CSS" tags, Unicode, few data entries that were not food recipes ("nutrient-calculator", "cnc--nyancat--food-mold--nyancake"). After removing the above-mentioned aberrations, in the next step, we used NLTK toolkit\cite{loper2002nltk} to process the out-of-dictionary vocabulary. Most of the out-of-dictionary vocabulary was found to be some form of a composite of in-dictionary vocabulary or vocabulary with numbers or measurement units. The next step of the preprocessing algorithm separates the in-dictionary words wrapped as out-of-dictionary words.  This process provided us with a much cleaner version of the data.

\section{Prior Scoring Functions}

\paragraph{Impatient Reader}\cite{hermann2015teaching} is an attention-based model that recurrently uses attention over the context for each question except the location containing the “@ placeholder”. This attention allows the model to accumulate information recurrently from the context as it sees each question embeddings. It outputs a final joint embedding for the answer prediction. This embedding is used to compute a compatibility score for each choice using a cosine similarity function in the feature space. The attention over context and question is computed on the output of an LSTM. The answer choices are also encoded using an LSTM with a similar architecture.

\paragraph{BiDAF} \cite{seo2016bidirectional} is abbreviated for "Bi-Directional Attentional Flow",  as the name suggests, it employs a bi-directional attention flow mechanism between the context, representation of the question images, and each candidate choice representation to learn temporal matching. We base our prediction on the best-matched candidate. Originally it was proposed as a span-selection model from the input context. Here, we adapt it to work in for visual tasks in multimodal setting.

\section{Experimental Results}

The additional visual cloze results on the remaining three pairwise combination of LM and VM hold true to our analysis in the main paper. The three tuple shown in the plots are: (Word2Vec, \vit), (BERT, ResNet--50), (BERT, \vit). We even see the impact of \ck{-3} in all three plots on the meta dataset as compared to it effect in the case of (Word2Vec, ResNet--50). In the case of coherence, we study the impacts using Word2Vec and ResNet-50, where \ck{-1} and \ck{-3} clearly have larger impact on model performance compared to \ck{-2}.


\begin{figure*}[htbp]
    \begin{center}
        \includegraphics[width=0.99\linewidth]{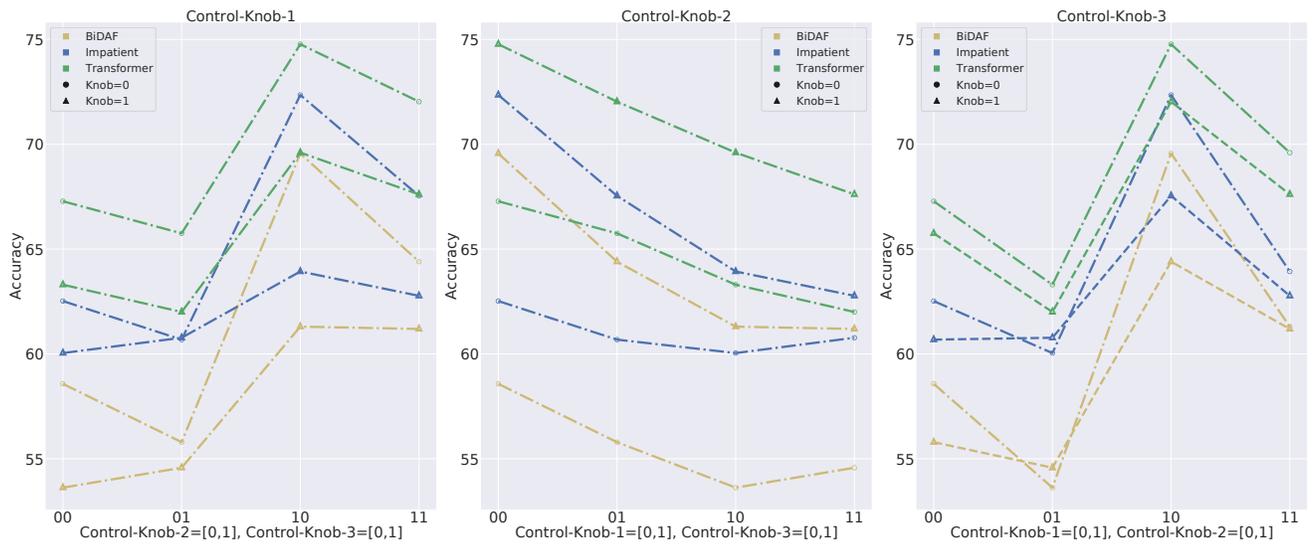}
    \end{center}
    \caption{\textbf{Visual Cloze: } Impact of \cks{} by comparing the performance of different scoring functions (adapted baseline and transformer) for each knob setting. LM is set to Word2Vec and VM is set to \vit. Starting from left we plot performance of \ck{-2} and \ck{-3} for all combination of control setting by fixing \ck{-1}. We do the same for \ck{-2} and \ck{-3} in the center and right figure respectively.} 
    \label{fig:impact_knobs1}
    \vspace{-1.3em}
\end{figure*}

\begin{figure*}[htbp]
    \begin{center}
        \includegraphics[width=0.99\linewidth]{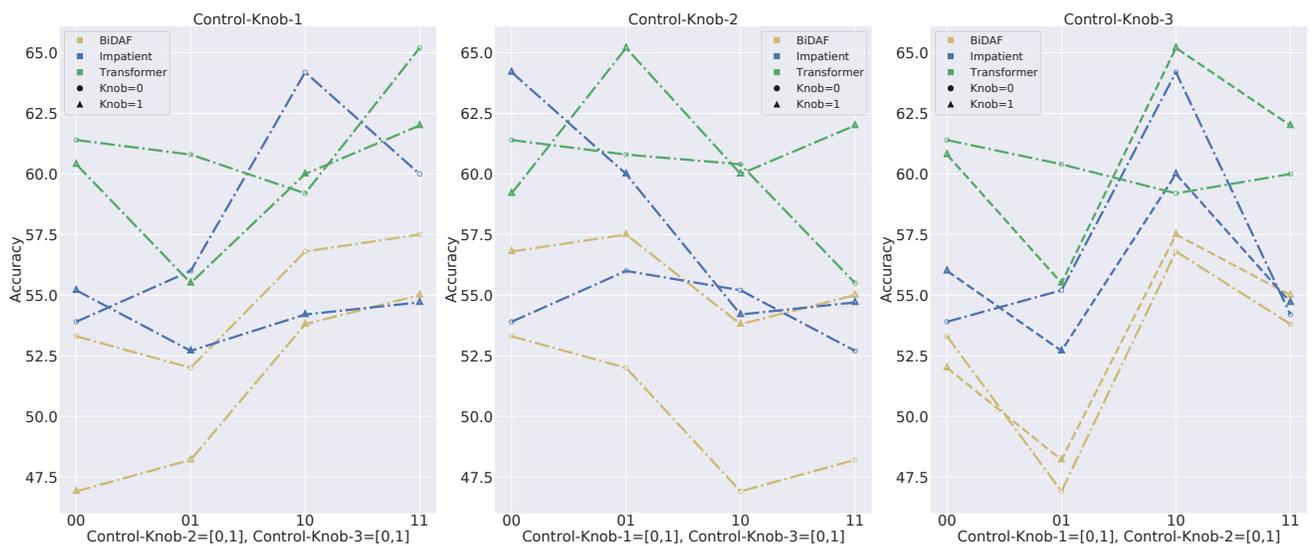}
    \end{center}
    \caption{\textbf{Visual Cloze: } Impact of \cks{} by comparing the performance of different scoring functions (adapted baseline and transformer) for each knob setting. LM is set to BERT and VM is set to ResNet--50. Starting from left we plot performance of \ck{-2} and \ck{-3} for all combination of control setting by fixing \ck{-1}. We do the same for \ck{-2} and \ck{-3} in the center and right figure respectively.}
    \label{fig:impact_knobs2}
    \vspace{-1.3em}
\end{figure*}

\begin{figure*}[htbp]
    \begin{center}
        \includegraphics[width=0.99\linewidth]{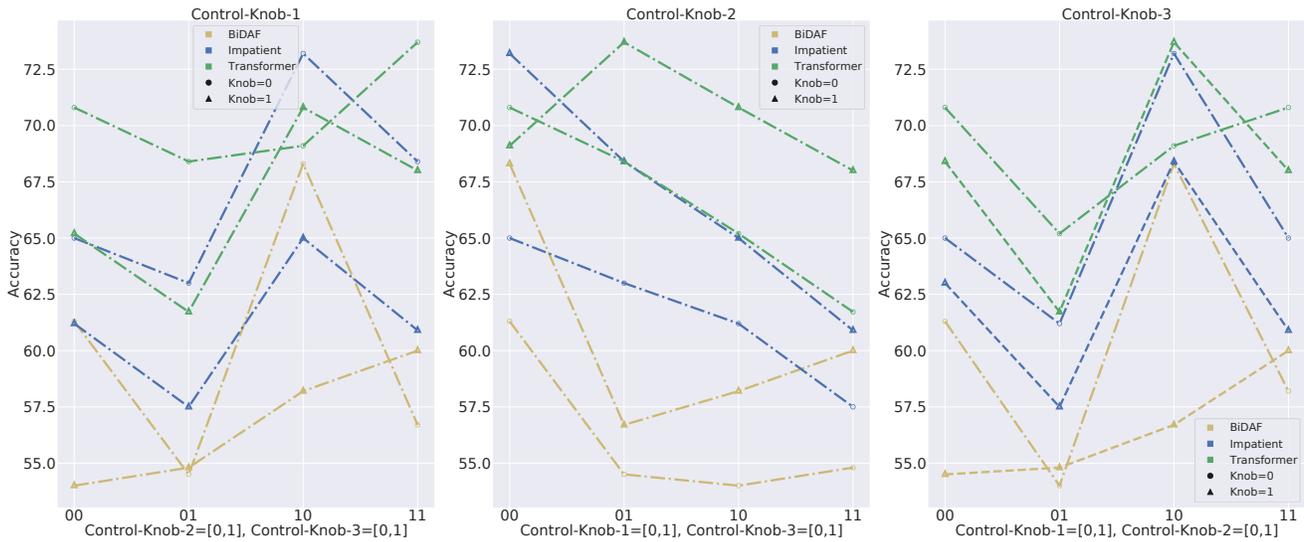}
    \end{center}
    \caption{\textbf{Visual Cloze: } Impact of \cks{} by comparing the performance of different scoring functions (adapted baseline and transformer) for each knob setting. LM is set to Bert and VM is set to \vit. Starting from left we plot performance of \ck{-2} and \ck{-3} for all combination of control setting by fixing \ck{-1}. We do the same for \ck{-2} and \ck{-3} in the center and right figure respectively.}
    \label{fig:impact_knobs3}
    \vspace{-1.3em}
\end{figure*}

\begin{figure*}[htbp]
    \begin{center}
        \includegraphics[width=0.99\linewidth]{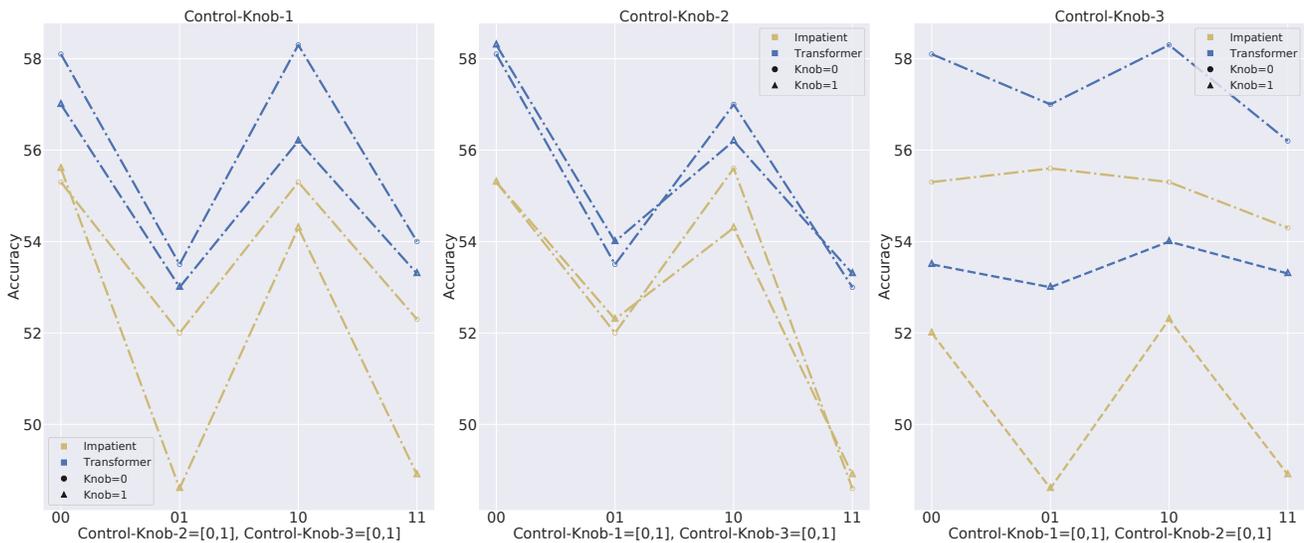}
    \end{center}
    \caption{\textbf{Visual Coherence: } Impact of \cks{} by comparing the performance of different scoring functions (adapted baseline and transformer) for each knob setting. LM is set to Word2Vec and VM is set to ResNet--50. Starting from left we plot performance of \ck{-2} and \ck{-3} for all combination of control setting by fixing \ck{-1}. We do the same for \ck{-2} and \ck{-3} in the center and right figure respectively.}
    \label{fig:impact_knobs4}
    \vspace{-1.3em}
\end{figure*}

\begin{figure*}[htbp]
    \begin{center}
        \includegraphics[width=0.99\linewidth]{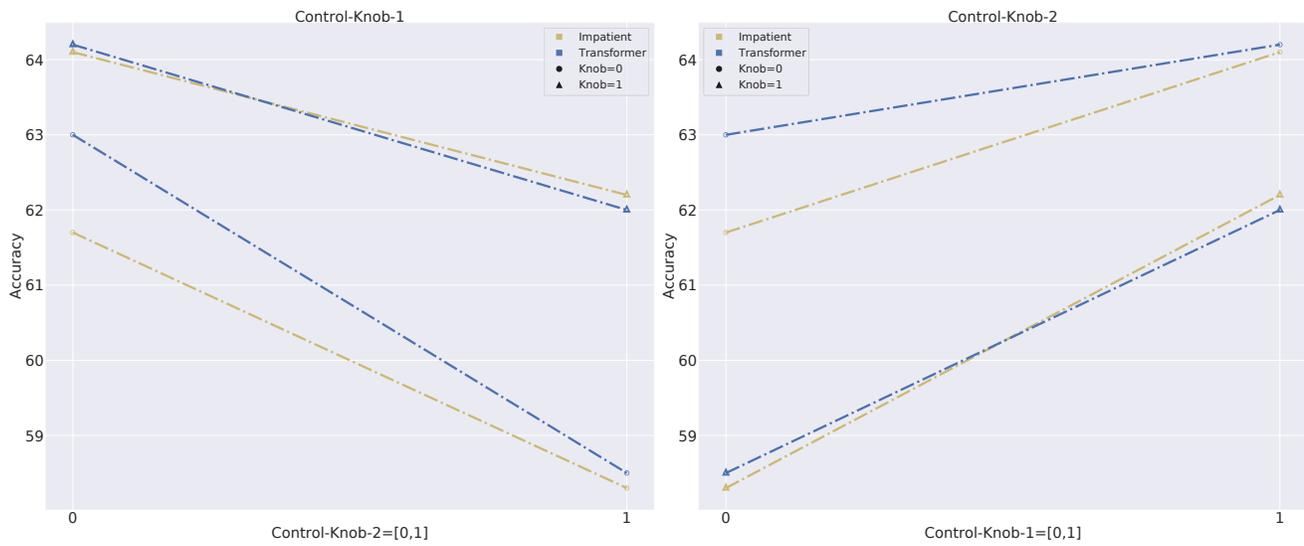}
    \end{center}
    \caption{\textbf{Visual Ordering: } Impact of \cks{} by comparing the performance of different scoring functions (adapted baseline and transformer) for each knob setting. LM is set to Word2Vec and VM is set to ResNet--50. For each \ck{} we fix one and plot the other, as show in the figure side by side.}
    \label{fig:impact_knobs5}
    \vspace{-1.3em}
\end{figure*}





%% file: main.bbl
\begin{thebibliography}{10}\itemsep=-1pt

\bibitem{amac2019procedural}
Mustafa~Sercan Amac, Semih Yagcioglu, Aykut Erdem, and Erkut Erdem.
\newblock Procedural reasoning networks for understanding multimodal
  procedures.
\newblock In {\em Proceedings of the 23rd Conference on Computational Natural
  Language Learning (CoNLL)}, pages 441--451, 2019.

\bibitem{antol2015vqa}
Stanislaw Antol, Aishwarya Agrawal, Jiasen Lu, Margaret Mitchell, Dhruv Batra,
  C Lawrence~Zitnick, and Devi Parikh.
\newblock Vqa: Visual question answering.
\newblock In {\em Proceedings of the IEEE international conference on computer
  vision}, pages 2425--2433, 2015.

\bibitem{brown2020language}
Tom~B Brown, Benjamin Mann, Nick Ryder, Melanie Subbiah, Jared Kaplan, Prafulla
  Dhariwal, Arvind Neelakantan, Pranav Shyam, Girish Sastry, Amanda Askell,
  et~al.
\newblock Language models are few-shot learners.
\newblock {\em arXiv preprint arXiv:2005.14165}, 2020.

\bibitem{cai2017pay}
Zheng Cai, Lifu Tu, and Kevin Gimpel.
\newblock Pay attention to the ending: Strong neural baselines for the roc
  story cloze task.
\newblock In {\em Proceedings of the 55th Annual Meeting of the Association for
  Computational Linguistics (Volume 2: Short Papers)}, pages 616--622, 2017.

\bibitem{chen2016thorough}
Danqi Chen, Jason Bolton, and Christopher~D Manning.
\newblock A thorough examination of the cnn/daily mail reading comprehension
  task.
\newblock In {\em Proceedings of the 54th Annual Meeting of the Association for
  Computational Linguistics (Volume 1: Long Papers)}, pages 2358--2367, 2016.

\bibitem{dasgupta2018evaluating}
Ishita Dasgupta, Demi Guo, Andreas Stuhlm{\"u}ller, Samuel~J Gershman, and
  Noah~D Goodman.
\newblock Evaluating compositionality in sentence embeddings.
\newblock {\em arXiv preprint arXiv:1802.04302}, 2018.

\bibitem{devlinetal2019bert}
Jacob Devlin, Ming-Wei Chang, Kenton Lee, and Kristina Toutanova.
\newblock {BERT}: Pre-training of deep bidirectional transformers for language
  understanding.
\newblock In {\em Proceedings of the 2019 Conference of the North {A}merican
  Chapter of the Association for Computational Linguistics: Human Language
  Technologies, Volume 1 (Long and Short Papers)}, pages 4171--4186,
  Minneapolis, Minnesota, June 2019. Association for Computational Linguistics.

\bibitem{gardner2019making}
Matt Gardner, Jonathan Berant, Hannaneh Hajishirzi, Alon Talmor, and Sewon Min.
\newblock On making reading comprehension more comprehensive.
\newblock In {\em Proceedings of the 2nd Workshop on Machine Reading for
  Question Answering}, pages 105--112, 2019.

\bibitem{gururangan2018annotation}
Suchin Gururangan, Swabha Swayamdipta, Omer Levy, Roy Schwartz, Samuel~R
  Bowman, and Noah~A Smith.
\newblock Annotation artifacts in natural language inference data.
\newblock {\em arXiv preprint arXiv:1803.02324}, 2018.

\bibitem{hermann2015teaching}
Karl~Moritz Hermann, Tomas Kocisky, Edward Grefenstette, Lasse Espeholt, Will
  Kay, Mustafa Suleyman, and Phil Blunsom.
\newblock Teaching machines to read and comprehend.
\newblock In {\em Advances in neural information processing systems}, pages
  1693--1701, 2015.

\bibitem{hill2016goldilocks}
Felix Hill, Antoine Bordes, Sumit Chopra, and Jason Weston.
\newblock The goldilocks principle: Reading children's books with explicit
  memory representations.
\newblock {\em International Conference on Learning Representations (ICLR)},
  2016.

\bibitem{hirschman1999deep}
Lynette Hirschman, Marc Light, Eric Breck, and John~D Burger.
\newblock Deep read: A reading comprehension system.
\newblock In {\em Proceedings of the 37th annual meeting of the Association for
  Computational Linguistics}, pages 325--332, 1999.

\bibitem{iyyer2017amazing}
Mohit Iyyer, Varun Manjunatha, Anupam Guha, Yogarshi Vyas, Jordan Boyd-Graber,
  Hal Daume, and Larry~S Davis.
\newblock The amazing mysteries of the gutter: Drawing inferences between
  panels in comic book narratives.
\newblock In {\em Proceedings of the IEEE Conference on Computer Vision and
  Pattern Recognition}, pages 7186--7195, 2017.

\bibitem{kaushik2018much}
Divyansh Kaushik and Zachary~C Lipton.
\newblock How much reading does reading comprehension require? a critical
  investigation of popular benchmarks.
\newblock In {\em Proceedings of the 2018 Conference on Empirical Methods in
  Natural Language Processing}, pages 5010--5015, 2018.

\bibitem{kembhavi2016diagram}
Aniruddha Kembhavi, Mike Salvato, Eric Kolve, Minjoon Seo, Hannaneh Hajishirzi,
  and Ali Farhadi.
\newblock A diagram is worth a dozen images.
\newblock In {\em European Conference on Computer Vision}, pages 235--251.
  Springer, 2016.

\bibitem{kingma2014adam}
Diederik~P Kingma and Jimmy Ba.
\newblock Adam: A method for stochastic optimization.
\newblock {\em arXiv preprint arXiv:1412.6980}, 2014.

\bibitem{liu2020multi}
Ao Liu, Shuai Yuan, Chenbin Zhang, Congjian Luo, Yaqing Liao, Kun Bai, and
  Zenglin Xu.
\newblock Multi-level multimodal transformer network for multimodal recipe
  comprehension.
\newblock In {\em Proceedings of the 43rd International ACM SIGIR Conference on
  Research and Development in Information Retrieval}, pages 1781--1784, 2020.

\bibitem{liu2019neural}
Shanshan Liu, Xin Zhang, Sheng Zhang, Hui Wang, and Weiming Zhang.
\newblock Neural machine reading comprehension: Methods and trends.
\newblock {\em Applied Sciences}, 9(18):3698, 2019.

\bibitem{loper2002nltk}
Edward Loper and Steven Bird.
\newblock Nltk: The natural language toolkit.
\newblock {\em arXiv preprint cs/0205028}, 2002.

\bibitem{mccoyetal2019right}
Tom McCoy, Ellie Pavlick, and Tal Linzen.
\newblock Right for the wrong reasons: Diagnosing syntactic heuristics in
  natural language inference.
\newblock In {\em Proceedings of the 57th Annual Meeting of the Association for
  Computational Linguistics}, pages 3428--3448, Florence, Italy, July 2019.
  Association for Computational Linguistics.

\bibitem{mikolov2013efficient}
Tomas Mikolov, Kai Chen, Greg Corrado, and Jeffrey Dean.
\newblock Efficient estimation of word representations in vector space.
\newblock {\em arXiv preprint arXiv:1301.3781}, 2013.

\bibitem{niven2019probing}
Timothy Niven and Hung-Yu Kao.
\newblock Probing neural network comprehension of natural language arguments.
\newblock In {\em Proceedings of the 57th Annual Meeting of the Association for
  Computational Linguistics}, pages 4658--4664, 2019.

\bibitem{poliaketal2018hypothesis}
Adam Poliak, Jason Naradowsky, Aparajita Haldar, Rachel Rudinger, and Benjamin
  Van~Durme.
\newblock Hypothesis only baselines in natural language inference.
\newblock In {\em Proceedings of the Seventh Joint Conference on Lexical and
  Computational Semantics}, pages 180--191, New Orleans, Louisiana, June 2018.
  Association for Computational Linguistics.

\bibitem{radford2021learning}
Alec Radford, Jong~Wook Kim, Chris Hallacy, Aditya Ramesh, Gabriel Goh,
  Sandhini Agarwal, Girish Sastry, Amanda Askell, Pamela Mishkin, Jack Clark,
  et~al.
\newblock Learning transferable visual models from natural language
  supervision.
\newblock {\em arXiv preprint arXiv:2103.00020}, 2021.

\bibitem{rajpurkar2016squad}
Pranav Rajpurkar, Jian Zhang, Konstantin Lopyrev, and Percy Liang.
\newblock Squad: 100,000+ questions for machine comprehension of text.
\newblock In {\em Proceedings of the 2016 Conference on Empirical Methods in
  Natural Language Processing}, pages 2383--2392, 2016.

\bibitem{reimers-2019-sentence-bert}
Nils Reimers and Iryna Gurevych.
\newblock Sentence-bert: Sentence embeddings using siamese bert-networks.
\newblock In {\em Proceedings of the 2019 Conference on Empirical Methods in
  Natural Language Processing}. Association for Computational Linguistics, 11
  2019.

\bibitem{richardson2013mctest}
Matthew Richardson, Christopher~JC Burges, and Erin Renshaw.
\newblock Mctest: A challenge dataset for the open-domain machine comprehension
  of text.
\newblock In {\em Proceedings of the 2013 Conference on Empirical Methods in
  Natural Language Processing}, pages 193--203, 2013.

\bibitem{sahu2021comprehension}
Pritish Sahu, Michael Cogswell, Sara Rutherford-Quach, and Ajay Divakaran.
\newblock Comprehension based question answering using bloom's taxonomy.
\newblock {\em arXiv preprint arXiv:2106.04653}, 2021.

\bibitem{sahu2021towards}
Pritish Sahu, Karan Sikka, and Ajay Divakaran.
\newblock Towards solving multimodal comprehension.
\newblock {\em arXiv preprint arXiv:2104.10139}, 2021.

\bibitem{schwartz2017effect}
Roy Schwartz, Maarten Sap, Ioannis Konstas, Leila Zilles, Yejin Choi, and
  Noah~A Smith.
\newblock The effect of different writing tasks on linguistic style: A case
  study of the roc story cloze task.
\newblock In {\em Proceedings of the 21st Conference on Computational Natural
  Language Learning (CoNLL 2017)}, pages 15--25, 2017.

\bibitem{seo2016bidirectional}
Minjoon Seo, Aniruddha Kembhavi, Ali Farhadi, and Hannaneh Hajishirzi.
\newblock Bidirectional attention flow for machine comprehension.
\newblock {\em International Conference on Learning Representations (ICLR)},
  2017a.

\bibitem{tapaswi2016movieqa}
M Tapaswi, Y Zhu, R Stiefelhagen, A Torralba, R Urtasun, and S Fidler.
\newblock Movieqa: Understanding stories in movies through question-answering.
\newblock In {\em 2016 IEEE Conference on Computer Vision and Pattern
  Recognition (CVPR)}, pages 4631--4640, 2016.

\bibitem{yagcioglu2018recipeqa}
Semih Yagcioglu, Aykut Erdem, Erkut Erdem, and Nazli Ikizler-Cinbis.
\newblock Recipeqa: A challenge dataset for multimodal comprehension of cooking
  recipes.
\newblock In {\em Proceedings of the 2018 Conference on Empirical Methods in
  Natural Language Processing}, pages 1358--1368, 2018.

\bibitem{zeng2020survey}
Changchang Zeng, Shaobo Li, Qin Li, Jie Hu, and Jianjun Hu.
\newblock A survey on machine reading comprehension—tasks, evaluation metrics
  and benchmark datasets.
\newblock {\em Applied Sciences}, 10(21):7640, 2020.

\end{thebibliography}
